\documentclass[final]{cvpr}

\usepackage{times}
\usepackage{epsfig}
\usepackage{graphicx}
\usepackage{amsmath}
\usepackage{amssymb}

\usepackage{subfigure}
\usepackage{floatrow}
\newfloatcommand{capbtabbox}{table}[][\FBwidth]

\usepackage[pagebackref=true,breaklinks=true,colorlinks,bookmarks=false]{hyperref}

\def\cvprPaperID{3115} 
\def\confYear{CVPR 2021}

\begin{document}

\title{Dynamic Region-Aware Convolution}

\author{Jin Chen\thanks{Equal contribution. This work is supported by The National Key Research and Development Program of China (No. 2017YFA0700800) and Beijing Academy of Artificial Intelligence (BAAI).}, Xijun Wang\footnotemark[1], Zichao Guo, Xiangyu Zhang, Jian Sun\\
MEGVII Technology\\
{\tt\small \{chenjin, wangxijun, guozichao, zhangxiangyu, sunjian\}@megvii.com}
}

\maketitle

\begin{abstract}
  We propose a new convolution called Dynamic Region-Aware Convolution (DRConv), which can automatically assign multiple filters to corresponding spatial regions where features have similar representation. In this way, DRConv outperforms standard convolution in modeling semantic variations. Standard convolutional layer can increase the number of filers to extract more visual elements but results in high computational cost. More gracefully, our DRConv transfers the increasing channel-wise filters to spatial dimension with learnable instructor, which not only improve representation ability of convolution, but also maintains computational cost and the translation-invariance as standard convolution dose. DRConv is an effective and elegant method for handling complex and variable spatial information distribution. It can substitute standard convolution in any existing networks for its plug-and-play property, especially to power convolution layers in efficient networks. We evaluate DRConv on a wide range of models (MobileNet series, ShuffleNetV2, etc.) and tasks (Classification, Face Recognition, Detection and Segmentation). On ImageNet classification, DRConv-based ShuffleNetV2-0.5$\times$ achieves state-of-the-art performance of 67.1\% at 46M multiply-adds level with 6.3\% relative improvement.
 
\end{abstract}

\begin{figure}[h]
\begin{center}
\includegraphics[width=1.0\linewidth]{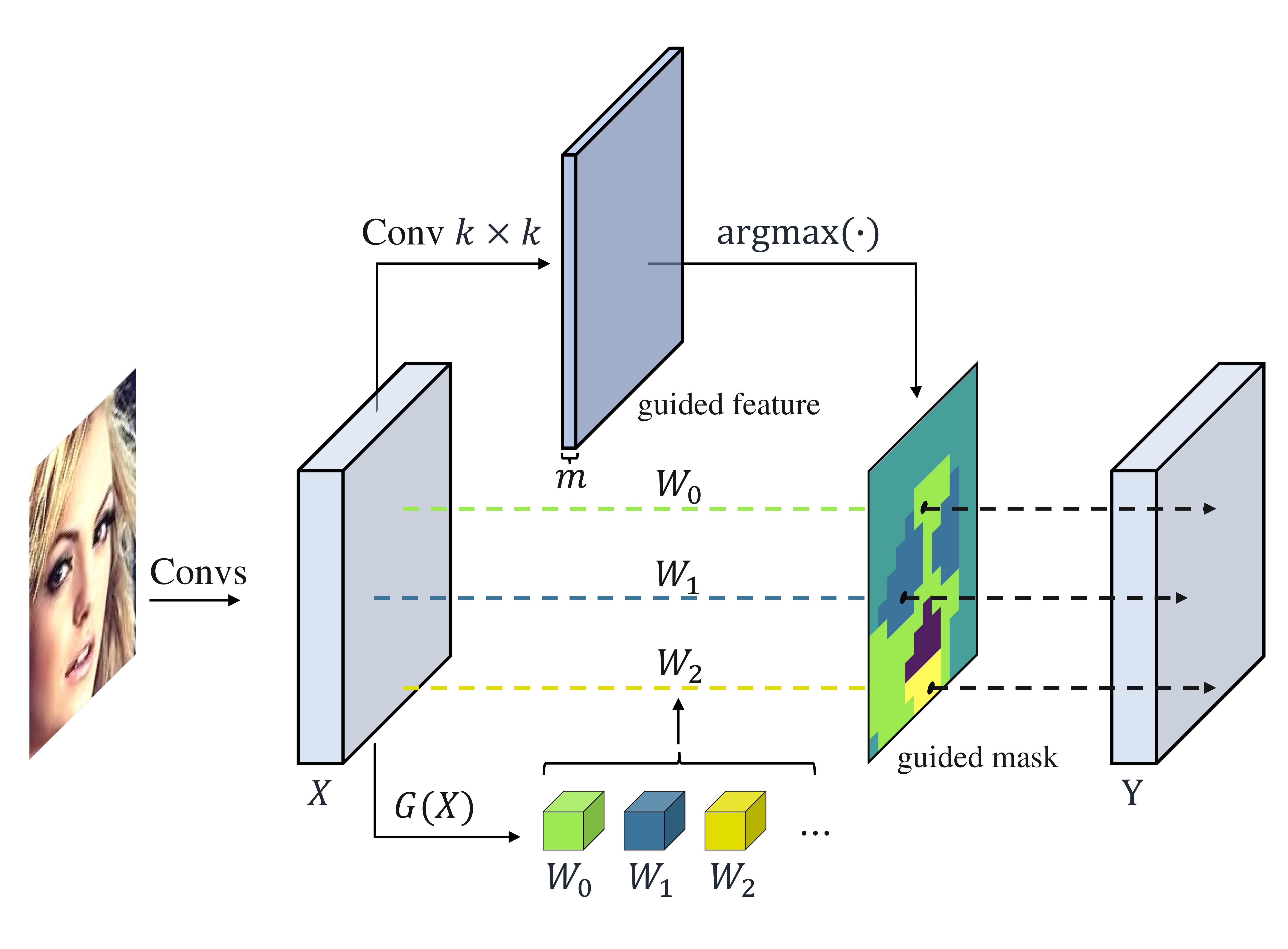}
\end{center}
\caption{Illustration of DRConv with kernel size $k \times k$ and region number $m$. We get guided feature from $X$ with standard $k \times k$ convolution and get $m$ filters from filter generator module $G(\cdot)$. The spatial dimension is divided into $m$ regions as guided mask shows. Every region has individual filter $W_{i}$ which is shared in this region and we execute $k \times k$ convolution with corresponding filter in these regions of $X$ to output $Y$.}
\label{Fig:DRConv}
\vspace*{-5mm}
\end{figure}

\section{Introduction} 
Benefiting from powerful representation ability, convolutional neural networks (CNNs) have made significant progress in image classification, face recognition, object detection and many other applications. The powerful representation ability of CNNs stems from that different filters are responsible for extracting information at different level of abstraction. However, current mainstream convolutional operations perform in filter sharing manner across spatial domain, so that more effective information can only be captured when these operations are applied repeatedly (e.g., increasing channels and depth with more filters). This repeating manner has several limitations. First, it is computationally inefficient. Second, it causes optimization difficulties that need to be carefully addressed \cite{he2016deep,wang2018non}. 

Different from the filter sharing methods, to model more visual elements, some studies focus on making use of the diversity of semantic information by using multiple filters in spatial dimension. \cite{gregor2010emergence,taigman2014deepface} came up with alternative convolutions to possess individual filter at each pixel in spatial dimension, and we collectively call them local convolution for convenience. Therefore, the feature of each position will be treated differently, which is more effective to extract the spatial feature than standard convolution. \cite{taigman2014deepface,sun2014deep,sun2014deep_2} have shown local convolution's power on face recognition task. Although local convolution doesn't increase the computation complexity compared with standard convolution, it has two fatal drawbacks. One is to bring a large amount of parameters, which is proportional to the spatial size. The other is that local convolution destroys translation-invariance, which is unfriendly to some tasks requiring translation-invariant features (e.g., local convolution doesn't work for classification task). Both of them make it hard to be widely used in neural networks. Besides, local convolution still shares filters across different samples, which makes it insensitive to the specific feature of each sample. For example, there are samples with different poses or viewpoints on face recognition and object detection tasks. Therefore, shared filters across different samples can not be effective to extract customized features. 

Considering above mentioned limitations, in this paper, we put forward a new convolution named Dynamic Region-Aware Convolution (DRConv), which can automatically assign filters to corresponding spatial-dimension regions with learnable instructor. As a consequence, DRConv has powerful semantic representation ability and perfectly maintains translation-invariance property. In detail, we design a learnable guided mask module to automatically generate the filters' region-sharing-pattern for each input image according to their own characteristic. The region-sharing-pattern means that we divide spatial dimension into several regions and only one filter is shared within each region. The filters for different samples and different regions are dynamically generated based on the corresponding input features, which is more effective to focus on their own vital characteristic. 

The structure of our DRConv is shown in Fig.~\ref{Fig:DRConv}. We apply standard convolution to generate guided feature from the input. According to the guided feature, the spatial dimension is divided into several regions. As can be seen, pixels with same color in the guided mask are attached to the same region. In each shared region, we apply filter generator module to produce a filter to execute 2D convolution operation. So the parameters needed to be optimized are mainly in filter generator module, and its amount has nothing to do with spatial size. Therefore, apart from significantly improving networks' performance, our DRConv can greatly reduce the amount of parameters compared with local convolution, and nearly doesn't increase the computation complexity compared with standard convolution.

To verify the effectiveness of our method, we conduct a series of empirical studies on several different tasks, including image classification, face recognition, object detection and segmentation by simply replacing standard convolution with our DRConv. The experimental results show that DRConv can achieve excellent performance on these tasks. We also offer adequate ablation studies for analyzing the effectiveness and robustness of our DRConv.

In brief, this work makes the following contributions,
\begin{enumerate}
    \item We present a new Dynamic Region-Aware Convolution, which not only has powerful semantic representation ability but also perfectly maintains translation-invariance property.
    \item We specially design the backward propagation process for learnable guided mask, so that our region-sharing-pattern is determined and updated according to the gradient of the overall task loss through backward propagation, which means our method can be optimized in an end-to-end manner. 
    \item Our DRConv can achieve excellent performance on image classification, face recognition, detection and segmentation tasks by simply replacing standard convolution without increasing much computation cost.
\end{enumerate}

\section{Related Work}
\label{sec:related_work}

We distinguish our work from other methods in term of spatial related work and dynamic mechanism. 
\\
\\
{\bf Spatial Related Convolution.} 
From the perspective of spatial related convolution design, the earliest enlightenment is local convolution. To effectively utilize the semantic information in image data, local convolution~\cite{gregor2010emergence} applies individual unshared filters to each pixel, which has great potential in tasks that don't require translation-invariance.
DeepFace~\cite{taigman2014deepface} and DeepID series~\cite{sun2014deep,sun2014deep_2} demonstrate the advantages of local convolution on face recognition task. These works illustrate that local distribution of spatial dimension is important.

On other tasks, such as detection, R-FCN~\cite{dai2016r} uses region-based fully convolutional networks to extract the local representations. It enlarges output channels to $3 \times 3$ times, and then selects corresponding subtensor in different channels to put together to form $3 \times 3$ blocks. On person re-identification, Sun et al.~\cite{sun2018beyond} applies part-based convolution to learn discriminative part-informed features, which can also be viewed as a kind of spatial related convolution. 

Besides the mentioned methods above, some studies try to change the spatial feature to better model the semantic variations. Spatial Transform Networks~\cite{mahajan2018exploring} learns transformation to warp the feature map but difficult to be trained. Jeon et al.~\cite{jeon2017active} introduces a convolution unit called active convolution unit (ACU), which produces unfixed shape because they can learn to any form through back propagation during training. ACU augments the sampling locations in the convolution with the learning offsets, and the offsets become static after training. Deformable Convolutional Networks~\cite{dai2017deformable} further makes the location offsets dynamic and then add the offsets to the regular grid sampling locations in the standard convolution. 

Compared with the mentioned studies above, our method adaptively divides the spatial dimension into several regions and shares one filter within each region. What's more, our design can maintain translation-invariance and extract more plentiful information.

{\bf Dynamic Mechanism.} 
With the prevalence of data dependency mechanism~\cite{allport1989visual,jaderberg2015spatial,vaswani2017attention} , which emphasizes to extract more customized feature~\cite{mahajan2018exploring}. Studies about dynamic mechanism have promoted many tasks to new state-of-the-art. Benefiting from data dependency mechanism, networks can flexibly adjust themselves, including the structure and parameters, to fit the diverse information automatically and improve representation ability of neural networks.

Some methods~\cite{chen2017sca,woo2018cbam} indicate that different regions in the spatial dimension are not equally important in representation learning. For instance, activation in important regions needs to be amplified so that it can play dominant role in the forward propagation. SKNet~\cite{li2019selective} designs an efficacious module to channel-wisely select suitable receptive fields on the basis of channel attention and achieves better performance. It dynamically restructures the networks for the sake of different receptive fields in dilated convolutions~\cite{yu2015multi,yu2017dilated}. In semantic segmentation, \cite{zhong2019squeeze} imposes a pixel-group attention to remedy the deficiency of spatial information in SENet and \cite{huang2019ccnet} builds a link between each pixel and its surrounding pixels to capture important information. Attention mechanism is designed to dynamically calibrate the information flow in the forward propagation by learnable method. 

From the aspect of dynamic weights, CondConv~\cite{yang2019condconv} obtains dynamic weights by dynamical linear combination of several weights. And the specialized convolution kernels for each sample are learned in a way similar to mixture of experts. In spatial domain, to handle object deformations, Deformable Kernels~\cite{gao2019deformable} directly resamples the original kernel space to adapt the effective receptive field (ERF). Local Relation Networks~\cite{hu2019local} adaptively determines aggregation weights for spatial dimension based on the compositional relationship of local pixel pairs. Non-local~\cite{wang2018non} operation computes the response at each position by weighted sum of the features at all positions, which can make it to capture long-range dependencies. 

Different from above dynamic methods, DRConv applies a dynamic guided mask to automatically determine the distribution of multiple filters so that it can handle variable distribution of spatial semantics.

\section{Our Approach}
\label{sec:Our_Approach}

The weight sharing mechanism inherently limits standard convolution to model semantic variations due to single filter's poor capacity. Therefore, standard convolution has to violently increase the number of filters across channel dimension to match more spatial visual elements, which is inefficient. Local convolution makes use of the diversity of spatial information but sacrifices translation-invariance. To deal with the above limitations once for all, we go a step further and propose a feasible solution called DRConv as Fig.~\ref{Fig:DRConv} shows, which not only increases the diversity of statistic 
by using more than one filter across spatial dimension, but also maintains translation-invariance for these positions having similar feature. 

\subsection{Dynamic Region-Aware Convolution}
We first briefly formulate standard convolution and basic local convolution, then transfer to DRConv. For convenience, we omit kernel size and stride of filters. The input of standard convolution can be denoted as $X \in \mathbb{R}^{U \times V \times C}$, where $U, V, C$ mean height, width and channel respectively. And $S \in \mathbb{R}^{U \times V}$ denotes spatial dimension, $Y \in \mathbb{R}^{U \times V \times O}$ for the output, and $W \in \mathbb{R}^{C}$ for standard convolution filters. For the $o$-th channel of output feature, the corresponding feature map is

\begin{equation}
    Y_{u, v, o} = \sum_{c=1}^{C} X_{u, v, c} \ast W^{(o)}_{c} \ \quad (u, v) \in S, 
\label{equ:standard_conv}
\end{equation}
where $ \ast $ is 2D convolution operation.

For basic local convolution, we use $W \in \mathbb{R}^{U \times V \times C}$ to denote the filters which don't share across spatial dimension. Therefore the $o$-th output feature map can be expressed as

\begin{equation}
    Y_{u, v, o} = \sum_{c=1}^{C} X_{u, v, c} \ast W^{(o)}_{u, v, c} \ \quad (u, v) \in S, 
\label{equ:local_conv}
\end{equation}
where $W^{(o)}_{u, v, c}$ represents individual unshared filter at pixel $(u,v)$ which is different from standard convolution.

Following above formulations, we define a guided mask $M = \{S_{0}, \cdots, S_{m-1}\}$ to represent the regions divided from spatial dimension, in which only one filter is shared in region $S_{t}$, $t \in [0, m-1]$. $M$ is learned from the input features according to the data dependency mechanism. We denote the filters of the regions as $W = [W_{0}, \cdots, W_{m-1}]$, where the filter $W_{t} \in \mathbb{R}^{C}$ is corresponding to the region $S_{t}$. The $o$-th channel of this layer's output feature map can be expressed as 

\begin{equation}
    Y_{u, v, g} = \sum_{c=1}^{C} X_{u, v, c} \ast W^{(o)}_{t, c} \quad (u, v) \in S_{t},
\label{equ:locally_shared_conv}
\end{equation}
where $W^{(o)}_{t, c}$ represents the c-th channel of the $W^{(o)}_{t}$ and $(u, v)$ is one of the points in region $S_{t}$. It needs to be noted that the point $(u, v)$ we use here is corresponding to the center of convolutional filter if the kernel size is larger than $1 \times 1$. That means a filter with kernel size $3 \times 3$ or $5 \times 5$ will extract features in adjacent regions on the border.

In general, our method is mainly divided into two steps. First, we use a learnable guided mask to divide the spatial features into several regions across spatial dimension. As Fig.~\ref{Fig:DRConv} shows, pixels with same color in the guided mask are attached to same region. And from the prospective of image semantics, the semantically similar features will be assigned to the same region.

Second, in each shared region, we use filter generator module to produce a customized filter to execute normal 2D convolution operation. The customized filter can be adjusted automatically according to the important characteristic of the input images. To better explain our method, we mainly introduce the following two modules: Learnable guided mask and Filter generator module. Learnable guided mask decides which filter will be assigned to which region. Filter generator module generates the corresponding filters which will be assigned to different regions.

\begin{figure*}[t]
\hspace{-0.5in}
\subfigure[Optimization of learnable guided mask]{
\centering
\includegraphics[scale=0.18]{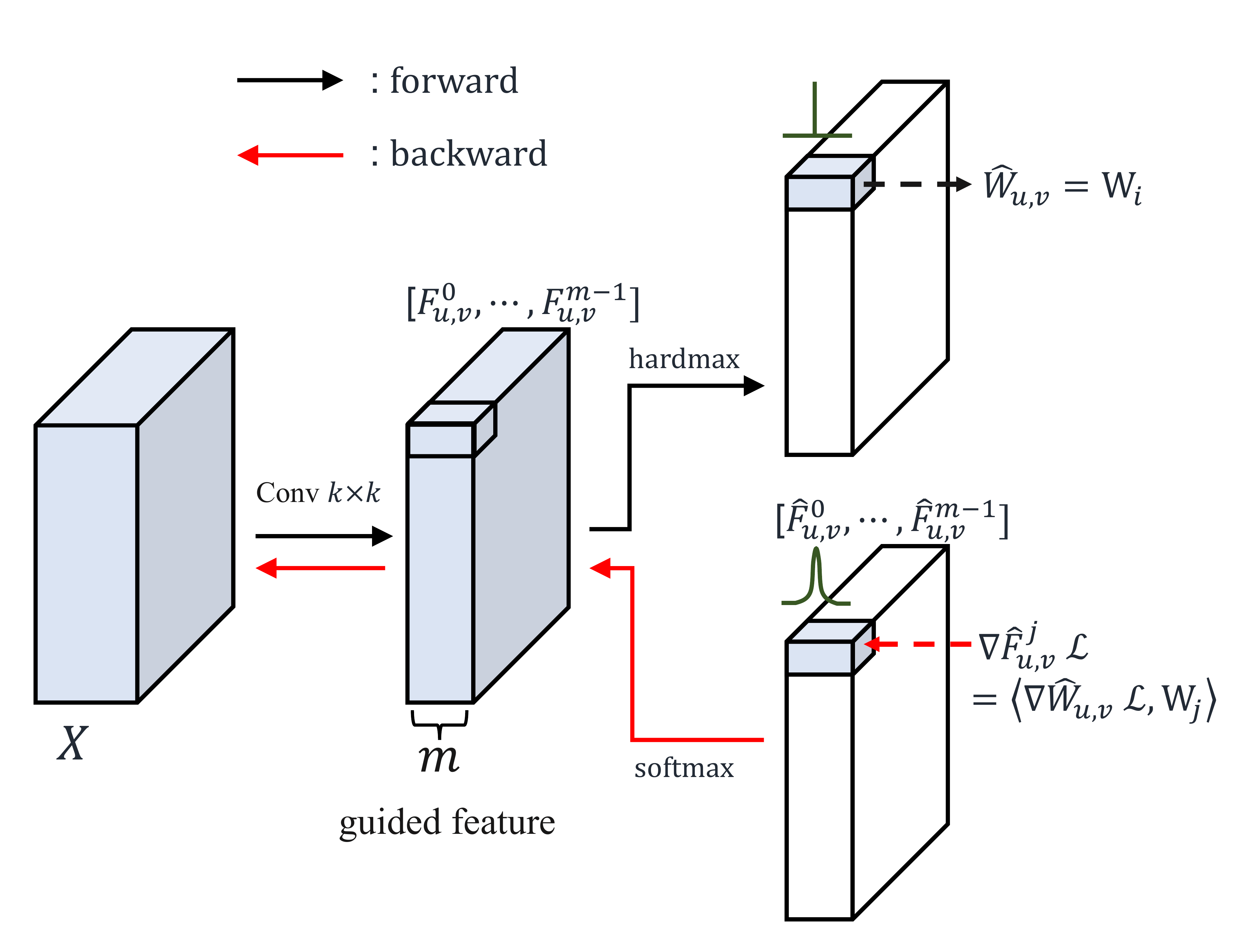}
}
\hspace{1.1in}
\subfigure[Filter generator module]{
\centering
\includegraphics[scale=0.18]{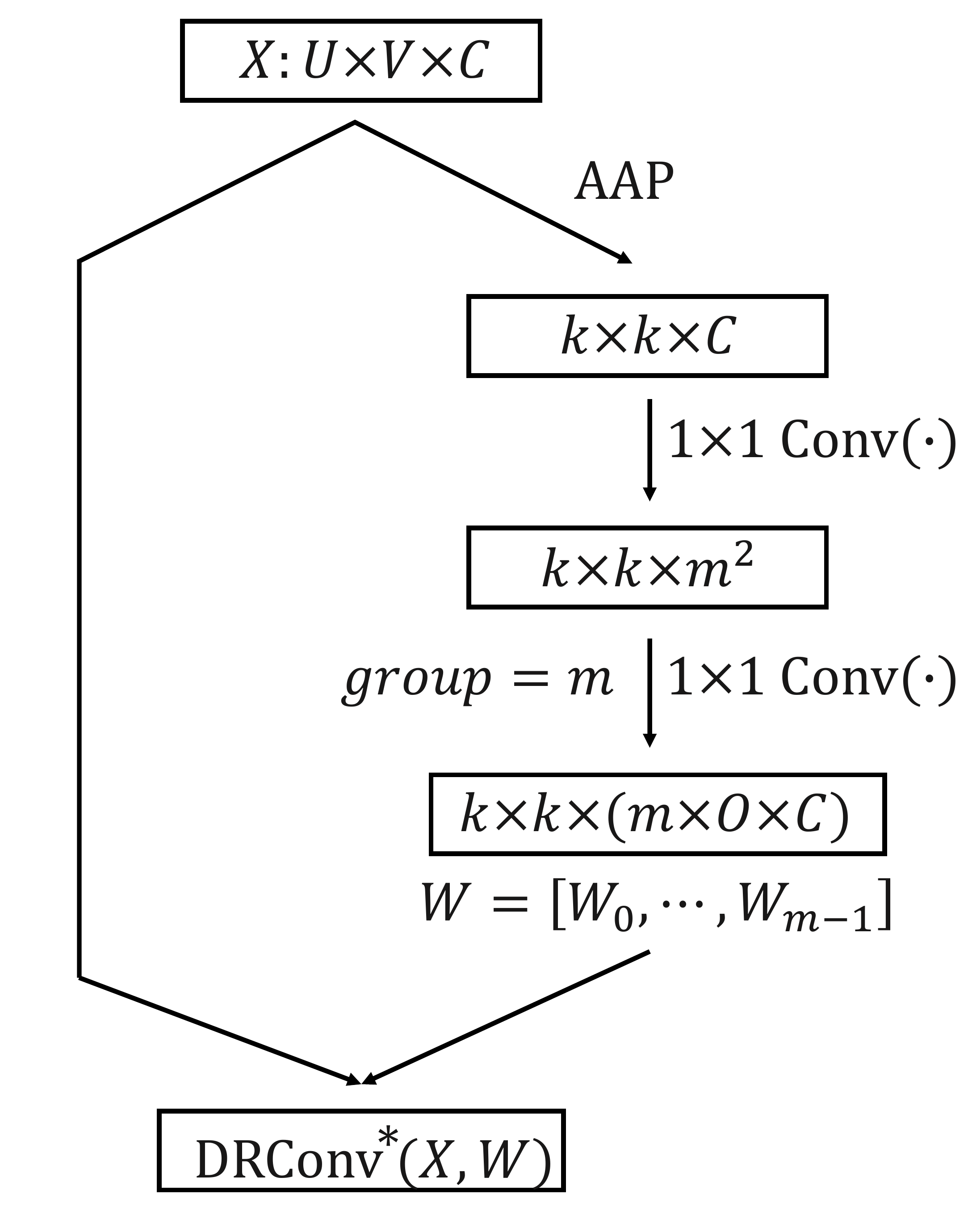}
}
\caption{(a) is the optimization process of learnable guided mask. Symbols are defined as Section \ref{subsec:Learnable_guided_mask}. We assign $\hat{W}_{u, v} = W_{i}$ when $F^{i}_{u, v}$ is the maximum across channel dimension. (b) is the architecture of DRConv for illustrating filter generator module. DRConv$^\ast$ denotes part of DRConv, AAP for adaptive average pooling and Conv($\cdot$) for standard convolution}
\label{Fig:Mask_and_Generate}
\vspace*{-4mm}
\end{figure*}

\subsection{Learnable guided mask}
\label{subsec:Learnable_guided_mask}
As one of the most important parts of our proposed DRConv, learnable guided mask decides the distribution (which filter will be assigned to which region) of filters across spatial dimension and is optimized by loss function. So that the filters can automatically adapt to the variance of spatial information for each input and the filter distribution will vary accordingly. In detail, as for a $k \times k$ DRConv with $m$ shared regions, we apply a $k \times k$ standard convolution to produce guided feature with $m$ channels ($k$ means kernel size). We use $F \in \mathbb{R}^{U \times V \times m}$ to denote guided feature, $M \in \mathbb{R}^{U \times V}$ for guided mask. For each position $(u, v)$ in spatial domain, we have
\begin{equation}
    M_{u, v} = argmax(\hat{F}^{0}_{u, v}, \hat{F}^{1}_{u, v}, \cdots, \hat{F}^{m-1}_{u, v}),
\label{equ:locally_shared_conv}
\end{equation}
where $argmax(\cdot)$ outputs the maximum value's index and $F_{u, v}$ denotes the vector of guided feature at position $(u, v)$ and has $m$ elements. So values in guided mask vary from $0$ to $m-1$ and indicate the index of filters which should be used in corresponding positions.

To make guided mask learnable, we must get the gradient for weights which produce guided feature. However, 
there is no gradient for guided feature, resulting in that the related parameters can not be updated. Therefore, we design approximate gradient for guided feature in an alternative way as shown in Fig.~\ref{Fig:Mask_and_Generate}(a). We will explain the forward and backward propagation in detail.
\\
\\
{\bf Forward propagation:}
Since we have the guided mask as Eq.(\ref{equ:locally_shared_conv}), we can get the filter $\hat{W}_{u, v}$ for each position $(u, v)$ as flowing:
\begin{equation}
    \hat{W}_{u, v} = W_{M_{u, v}} \quad M_{u, v}\in [0, m-1] = W*M_{u, v},
\label{equ:forward_select}
\end{equation}
where $W_{M_{u, v}}$ is one of the filters $[W_{0}, \cdots, W_{m-1}]$ generated by $G(\cdot)$ and $M_{u, v}$ is the index of the maximum across the channel dimension of guided feature $F$ at position $(u, v)$. In this way, $m$ filters will build corresponding relationship with all positions and the entire spatial pixels can be divided into $m$ groups. These pixels using the same filter are of similar context because a standard convolution with translation-invariance conveys their information to guided feature.
\\
\\
{\bf Backward propagation:}
As shown in Fig.~\ref{Fig:Mask_and_Generate}(a), we first introduce $\hat{F}$, which is the substitution of guided mask's one-hot-form(e.g., $M_{u, v}=2, m=5$, and $M_{u, v}$'s one-hot-form is $[0,0,1,0,0]$) in backward propagation,
\begin{equation}
    \hat{F}^{j}_{u, v} = \frac{e^{F^{j}_{u, v}}}{\sum_{n=0}^{m-1} e^{F^{n}_{u, v}}} \quad j\in [0, m-1],
\label{equ:softmax}
\end{equation}
Eq.(\ref{equ:softmax}) is $softmax(\cdot)$ function, which is applied to guided feature $F$ across channel dimension.
With $softmax$ operation, we expect $\hat{F}^{j}_{u, v}$ to approximate 0 or 1 as close as possible. As a result, the gap between $\hat{F}^{j}_{u, v}$ and guided mask's one-hot-form becomes very small. Moreover, $\hat{W}_{u, v}$  in Eq.(\ref{equ:forward_select}) can be viewed as the filters $[W_{0}, \cdots, W_{m-1}]$ multiplied by the one-hot-form of $M_{u, v}$ which can be approximated by $[\hat{F}^{0}_{u, v}, \cdots, \hat{F}^{m-1}_{u, v}]$. 
Then the gradient of $\hat{F}^{j}_{u, v}$ can be got by 
\\
\begin{equation}
    \bigtriangledown_{\hat{F}^{j}_{u, v}} \mathcal{L} = \langle \bigtriangledown_{\hat{W}_{u, v}} \mathcal{L}, W_{j} \rangle \quad j\in [0, m-1],
\label{equ:backward_select}
\end{equation}
\\
where $\langle,\rangle$ denotes dot product and $\bigtriangledown_{\cdot} \mathcal{L}$ means the tensor's gradient with respect to loss function. As Fig.~\ref{Fig:Mask_and_Generate}(a) shows, Eq.(\ref{equ:backward_select}) is the approximate backward propagation of Eq.(\ref{equ:forward_select}). 

\begin{equation}
    \bigtriangledown_{F_{u, v}} \mathcal{L} = \hat{F}_{u, v} \odot (\bigtriangledown_{\hat{F}_{u, v}} \mathcal{L} - \mathbf{1} \langle \hat{F}_{u, v}, \bigtriangledown_{\hat{F}_{u, v}} \mathcal{L} \rangle) ,
\label{equ:backward_softmax}
\end{equation}
where $\odot$ denotes element-by-element multiplication and Eq.(\ref{equ:backward_softmax}) is exactly the backward propagation of Eq.(\ref{equ:softmax}). If we don't design special backward propagation, SGD can't optimized relevant parameters because the function $argmax(\cdot)$ is non-differentiable and will stop the propagation of gradient. Therefore, $softmax(\cdot)$ function is used to be an approximate replacement of $argmax(\cdot)$ in backward propagation, which is differentiable and will minify the gap between the two function's outputs. More importantly, we can utilize it to transfer the gradient to guided feature so that guided mask can be optimized.

\subsection{Dynamic Filter: Filter generator module}
In our DRConv, multiple filters will be assigned to different regions. Filter generator module is designed to generate filters for these regions. Due to the diversity of characteristics among different images, shared filter across images is not effective enough to focus on their own characteristics. Such as images with different poses and viewpoints in face recognition and object detection tasks, which customized features are needed to focus on specific characteristic of each image.

Following the symbols we use above, we denote the input as $X \in \mathbb{R}^{U \times V \times C}$ and $G(\cdot)$ for filter generator module which mainly includes two convolution layers. The $m$ filters are denoted as $W = [W_{0}, \cdots, W_{m-1}]$ and each filter is only shared in one region $R_{t}$. As shown in Fig.~\ref{Fig:Mask_and_Generate}(b), to get $m$ filters with kernel size $k \times k$, we use adaptive average pooling to downsample $X$ to size $k \times k$. Then we apply two consecutive $1 \times 1$ convolution layers: the first uses $sigmoid(\cdot)$ as activation function and the second with $group = m$ doesn't use activation function. Filter generator module can enhance the ability of capturing specific characteristics of different images. As Fig.~\ref{Fig:Mask_and_Generate}(b) shows, the filters for convolution are predicted based on the feature of each sample respectively. So each filter can be adjusted automatically according to their own characteristic. 

\setlength{\tabcolsep}{4pt}
\begin{table*}[htp]
\begin{center}
\caption{Comparisons with state-of-the-art in terms of Top-1 classification accuracy (\%) on ImageNet. DRConv outperforms previous methods (global 2$\times$2 \& local 4$\times$4 means global $1\times1$ Deformable kernels with scope size 2$\times$2 and local $3\times3$ Deformable kernels with scope size 4$\times$4. $*$ means using fewer computational cost. MADDs refers to the number of multiply-adds operations)}
\label{table:main_results}
\begin{tabular}{cccc}
\hline\noalign{\smallskip}
Model & 
Setting &
MADDs & 
\begin{tabular}[c]{@{}l@{}}Top-1 ACC.(\%)\end{tabular} \\
\noalign{\smallskip}
\hline 
\noalign{\smallskip}
ShuffleNetV2 0.5$\times$     & baseline          & 42M               & 60.8 \\
CondConv-ShuffleNetV2 0.5$\times$   & 8 weights         & 43M               & 65.0 \\
DRConv-ShuffleNetV2 0.5$\times$* (Ours)    & 8 regions          & \textbf{39M}               & \textbf{64.9} \\
DRConv-ShuffleNetV2 0.5$\times$ (Ours)     & 8 regions          & 46M               & \textbf{67.1} \\ 
\hline
ShuffleNetV2 1$\times$              & baseline          & 147M               & 69.5 \\
CondConv-ShuffleNetV2 1$\times$     & 8 weights         & 152M               & 72.0 \\
DRConv-ShuffleNetV2 1$\times$* (Ours)      & 8 regions          & \textbf{109M}               & \textbf{72.2} \\ 
DRConv-ShuffleNetV2 1$\times$ (Ours)       & 8 regions          & 157M               & \textbf{73.1} \\ 
\hline
MobileNetV2~\cite{sandler2018mobilenetv2}                  & baseline          & 300M               & 72.0 \\
CondConv-MobileNetV2~\cite{yang2019condconv}         & 8 weights         & 329M               & 74.6 \\
DK-MobileNetV2~\cite{gao2019deformable}  & global 2$\times$2 \& local 4$\times$4  & 760M               & 74.8 \\
DRConv-MobileNetV2* (Ours)          & 8 regions          & \textbf{201M}               & \textbf{74.7} \\
DRConv-MobileNetV2 (Ours)          & 8 regions          & 328M               & \textbf{75.7} \\ \hline


MobileNetV1~\cite{howard2017mobilenets}                  & baseline          & 569M               & 70.6 \\
CondConv-MobileNetV1~\cite{yang2019condconv}         & 8 weights          & 600M               & 73.7 \\
DRConv-MobileNetV1* (Ours)           & 8 regions          & \textbf{344M}               & \textbf{74.4} \\ 
DRConv-MobileNetV1 (Ours)           & 8 regions          & 610M               & \textbf{75.5} \\ 
\hline
\end{tabular}
\end{center}
\vspace*{-5mm}
\end{table*}
\setlength{\tabcolsep}{1.4pt}

\section{Experiments}
In this section, we will prove the effectiveness of our proposed DRConv by embedding it into the existing popular neural networks including ShuffleNetV2~\cite{ma2018shufflenet} and MobileNet series~\cite{howard2017mobilenets,sandler2018mobilenetv2}. We compare DRConv with existing state-of-the-art on ImageNet~\cite{russakovsky2015imagenet}, MS1M-V2~\cite{guo2016ms}, and COCO in terms of image classification, face recognition, object detection and segmentation. Unless otherwise specified, all the experiments of DRConv are based on 8-learnable-region (i.e. $m=8$).

\subsection{Classification}
The ImageNet 2012 dataset~\cite{russakovsky2015imagenet} is a widely accepted and authoritative image classification dataset, consisting of $1.28$ million training images and 50k validation images from 1000 classes. Following the mainstream works, all the models are trained on the entire training dataset and evaluated by the single-crop top-1 validation set accuracy. And for both training and evaluation, the input image resolution is $224\times224$. The training setting follows \cite{ma2018shufflenet}, all models in our experiments are trained for 240 epochs, with learning rate which starts from $0.5$ and linearly reduces to $0$. 

To prove the effectiveness of DRConv, we compare DRConv with state-of-the-art methods including \cite{gao2019deformable,yang2019condconv}. The results are shown in Table~\ref{table:main_results}. In the first column, for example, CondConv-ShuffleNetV2 means that all $1\times1$ standard convolutions in ShuffleNetV2 are replaced by CondConv~\cite{yang2019condconv}. For DRConv-based model, we replace all the $1\times1$ standard convolutions in the backbone with DRConv. As can be seen, with comparable computational cost, DRConv-ShuffleNetV2 achieves 6.3\% and 3.6\% gain over ShuffleNetV2 for $0.5\times$ and $1\times$ scale respectively. DRConv-MobileNetV2 achieves a 3.7\% gain over MobileNetV2 and DRConv-MobileNetV1 achieves a 4.9\% gain over the baseline MobileNetV1. We also evaluate our approach by using fewer computational cost, and found that we still had an advantage over CondConv in using fewer calculations as shown in model tagged with $*$. These experimental results show that DRConv-based networks not only have a considerable improvement over the baselines, but also a great improvement over the state-of-the-art methods, demonstrating the effectiveness of our method.

As a basic of some other tasks, classification needs to extract as ample information as possible for predicting image's label because the large number of categories in ImageNet dataset. Traditional large networks can realize state-of-the-art due to their huge depth and width. As for efficient networks which are expected to be used in practice, they need to improve the efficiency of extracting useful information under the constraint of limited depth and width. Therefore, we design DRConv to augment representation capacity by making full use of the diversity of spatial information without extra computation cost. The multi-filter strategy for spatial information means it can match more information pattern. 

\setlength{\tabcolsep}{4pt}
\begin{table}[t]
\begin{center}
\caption{Results of DRConv on Megaface. ``ACC.'' refers to the rank-1 face identification accuracy with 1M distractors. (Training dataset: MS1M-V2)}
\label{table:face}
\begin{tabular}{ccc}
\hline\noalign{\smallskip}
Model & 
\begin{tabular}[c]{@{}l@{}}MADDs ($\times 10^6$)\end{tabular} & 
\begin{tabular}[c]{@{}l@{}} ACC.(\%)\end{tabular} \\
\noalign{\smallskip}
\hline 
\noalign{\smallskip}
MobileFaceNet                & 189               & 91.3 \\
Local-MobileFaceNet          & 189               & 94.9 \\
CondConv-MobileFaceNet       & 195               & 94.8 \\
DRConv-MobileFaceNet         & 201               & \textbf{96.2} \\
\hline
\end{tabular}
\end{center}
\vspace*{-5mm}
\end{table}
\setlength{\tabcolsep}{1.4pt}

\subsection{Face Recognition}
We use MobileFaceNet~\cite{chen2018mobilefacenets} as our backbone which has only 1M parameters and 189M MADDs with input size $112 \times 96$. To keep stability of training, we replace the Arcface loss~\cite{deng2019arcface} with AM-Softmax loss~\cite{wang2018additive} in our implementation. The dataset we use for training is MS1M-V2, which is introduced as a large-scale face dataset with 5.8M images from 85k celebrities. It is a semi-automatic refined version of the MS-Celeb-1M dataset~\cite{guo2016ms} which consists of 1M photos from 100k identities and has a large number of noisy image or wrong ID labels. The dataset we use for evaluation is MegaFace~\cite{kemelmacher2016megaface}, which includes 1M images of 60k identities as the gallery set and 100k images of 530 identities from FaceScrub as the probe set. Due to the same reason, it is also a refined version by manual clearing. 

{\bf Training and Evaluation:} We use SGD with momentum 0.9 to optimize the model and the batch size is 512. We train all the models for 420k iterations. The learning rate begins with 0.1 and is divided by 10 at 252k, 364k and 406k iterations. The setting of weight decay follows \cite{chen2018mobilefacenets}. For evaluation, we use face identification metric which refers to the rank-1 accuracy on MegaFace as the evaluation indicator.

\begin{figure*}[t]
\begin{center}
\includegraphics[width=0.9\linewidth]{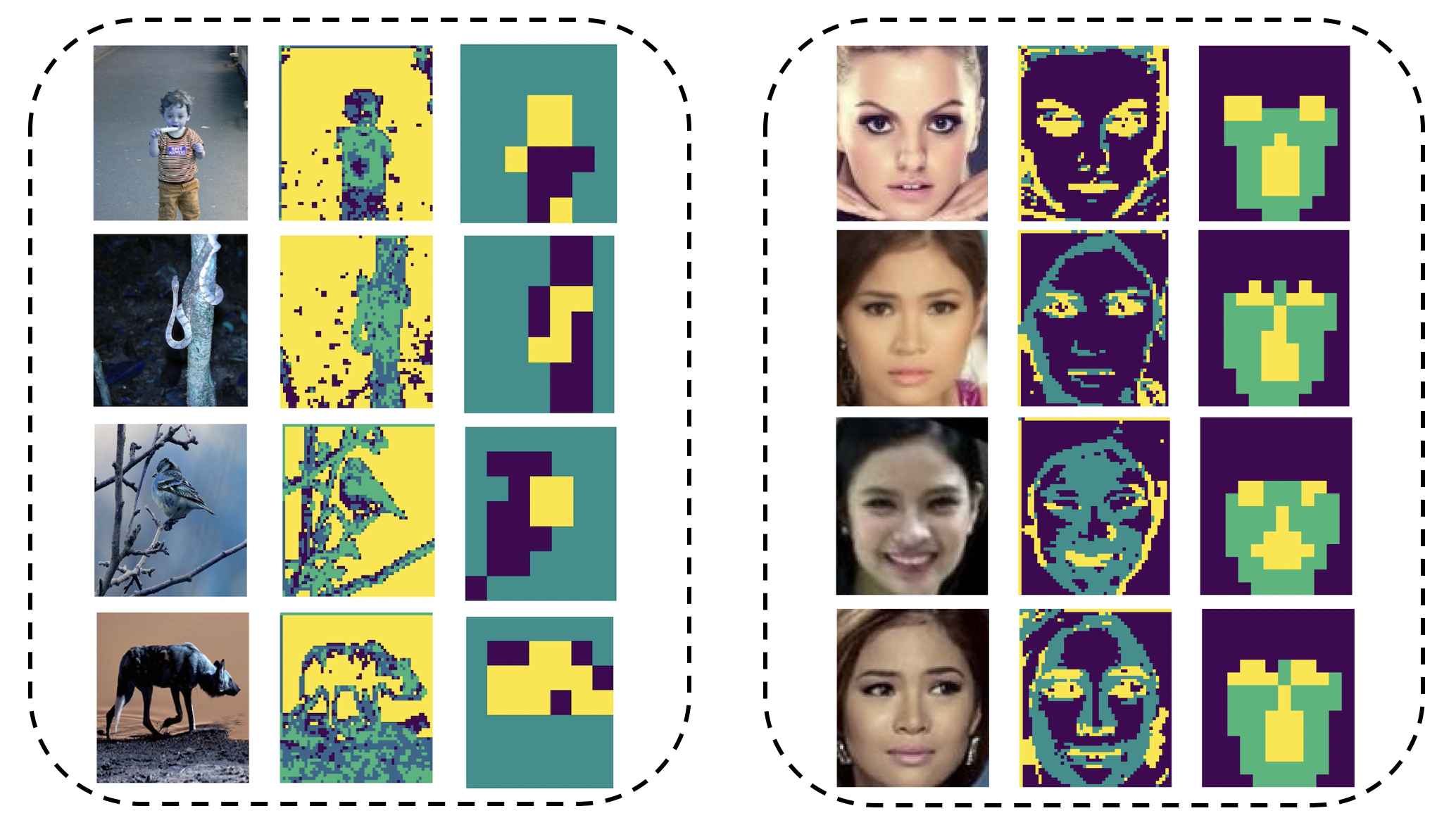}
\end{center}
\caption{Visualizations: guided mask of image on classification and face recognition. The first column represents the original images. The second and third columns represent visualizations of different layers' guided masks, respectively}
\label{Fig:visual}
\vspace*{-4mm}
\end{figure*}

In order to verify the effectiveness of our DRConv, we compare DRConv with several related methods. Based on the MobileFaceNet backbone, we simply replace $1 \times 1$ standard convolution in all bottleneck blocks with our DRConv.  As Table~\ref{table:face} shows, DRConv-MobileFaceNet outperforms the baseline by a margin of 4.9\%, and achieves a 1.4\% gain over CondConv. For further comparison, we choose local convolution which works for face recognition but needs huge amount of parameters. Under the limitation of device memory, we apply local convolution in the last three layers. DRConv-MobileFaceNet achieves 1.3\% higher accuracy than Local-MobileFaceNet(using local convolution in MobileFaceNet), further indicating the superiority of our proposed DRConv. Due to spatial stationarity of local statistics in face dataset, DRConv's guided mask module can learn clear semantic pattern. As shown in Fig.~\ref{Fig:visual}, the facial components appear in these guided masks.

\setlength{\tabcolsep}{4pt}
\begin{table*}
\begin{center}
\caption{Results of DRConv on COCO object detection and segmentation. ``R'' in 8R denotes region number. We replace $1\times1$ standard convolutions in the backbone of DetNAS-300M and only \emph{two-layer} in FPN of Mask R-CNN with DRConvs.}
\label{table:det_and_seg}
\begin{tabular}{c|ccc|ccc}
\hline 
Model         & ${\rm AP^{bbox}}$ & ${\rm AP^{bbox}_{50}}$    & ${\rm AP^{bbox}_{75}}$ & ${\rm AP^{mask}}$ & ${\rm AP^{mask}_{50}}$    & ${\rm AP^{mask}_{75}}$    \\
\hline
DetNAS-300M         &  36.6  &  57.4  &  39.3  & $\setminus$  &  $\setminus$  &  $\setminus$  \\
DRConv-DetNAS-300M 8R   &  \textbf{38.4}  &  \textbf{59.6}  &  \textbf{41.6}  & $\setminus$  &  $\setminus$  &  $\setminus$  \\ \hline
MaskRCNN            &  39.1  &  59.0  &  42.8  & 34.5  &  55.8  &  36.6  \\
DRConv-MaskRCNN 4R  &  39.8  &  60.3  &  43.3  & 35.3  &  57.1  &  37.4  \\ 
DRConv-MaskRCNN 8R  &  40.2  &  60.8  &  44.0  & 35.5  &  57.6  &  37.6  \\
DRConv-MaskRCNN 16R &  \textbf{40.3}  &  \textbf{61.2}  &  \textbf{44.2}  & \textbf{35.6}  &  \textbf{58.0}  &  \textbf{37.6}  \\  \hline
\end{tabular}
\end{center}
\vspace*{-4mm}
\end{table*}
\setlength{\tabcolsep}{1.4pt}

\begin{figure}
\begin{center}
\subfigure[DRConv-ShuffleNetV2]{
\includegraphics[scale=0.22]{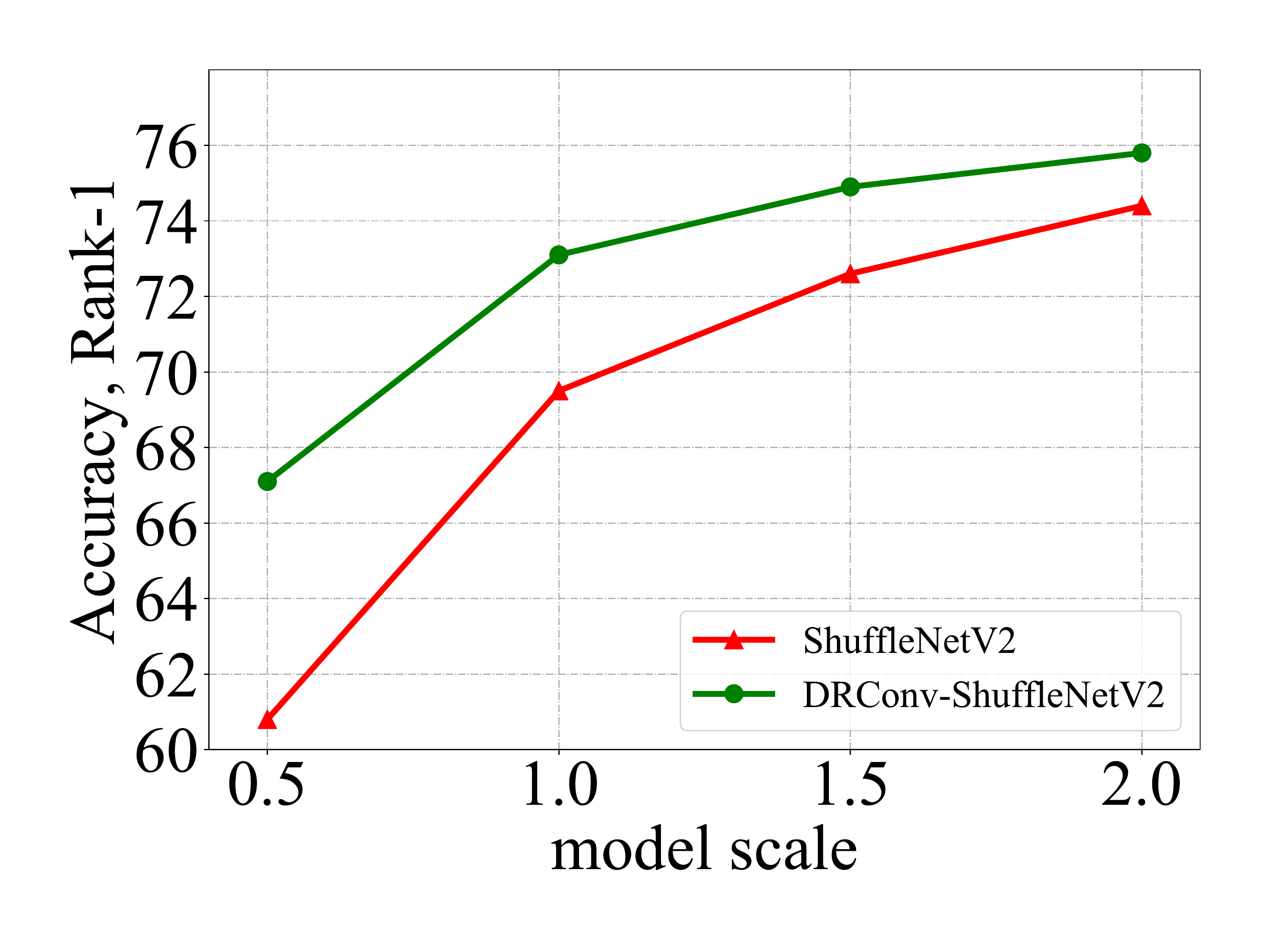}
}
\subfigure[DRConv-MobileNetV2]{
\includegraphics[scale=0.22]{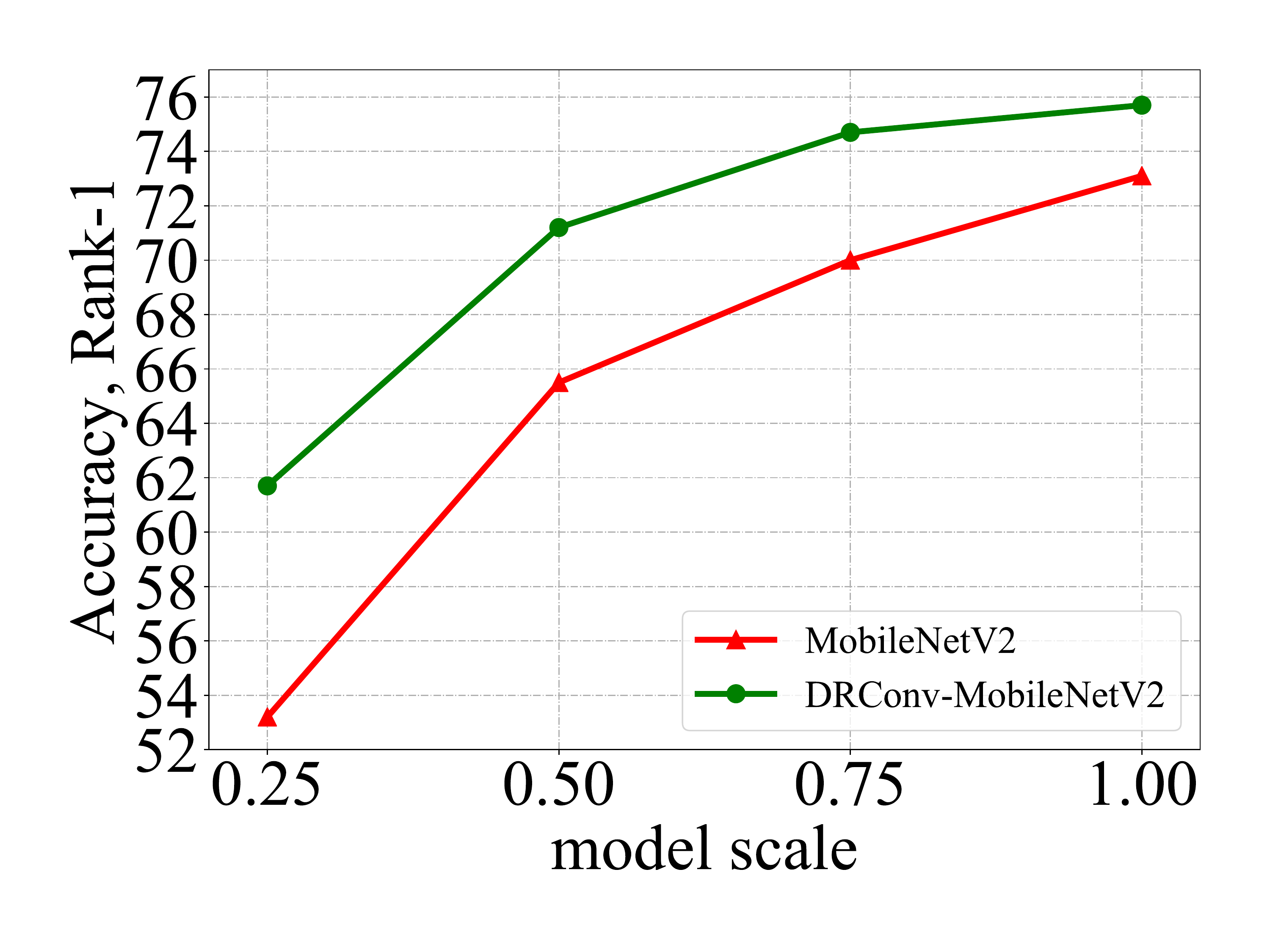}
}
\end{center}
\caption{Results of DRConv-ShuffleNetV2 and DRConv-MobileNetV2 on different model size. Small models will achieve higher gain}
\label{Fig:Comparision_DRConv_vs_Shuffle_and_Mobile}
\end{figure}

\subsection{COCO Object Detection and Segmentation}
We further evaluate the effectiveness of our DRConv on object detection and segmentation. We use the COCO dataset which consists of 80k train images and 40k val images. As in many previous works, we train on the union of 80k train images and a 35k subset of val images excluding 5k minival images, on which we evaluate our DRConv.

In experiments, we use DetNAS-300M~\cite{NIPS2019_8890} and Mask R-CNN~\cite{he2017mask} framework with FPN~\cite{lin2017feature} and a 4conv1fc box head as the basis to assess our method. Weights are initialized by the parameters of ClsNASNet~\cite{NIPS2019_8890} and ResNet50~\cite{he2016deep} respectively which are trained on the ImageNet dataset~\cite{russakovsky2015imagenet} and used as the feature extractor. In DetNAS-300M, training settings follow \cite{NIPS2019_8890}. In Mask R-CNN, the number of proposals in the head part for possible objects is set as 512. We train the detection and segmentation networks on 8 GPU with a batch size of 16 for 180k iterations. At beginning, we warm up the networks with factor $0.33$ for 500 iterations. During the training process, we use the learning rate 0.2 and decay the learning rate by 0.1 times at 120k, 140k and 150k iterations.

Our goal is to assess the effect when we replace $1\times1$ standard convolutions in the backbone of DetNAS-300M and only \emph{two-layer} in FPN of Mask R-CNN with DRConvs, so that any improvement on performance can be attributed to the effectiveness of our DRConv. In addition, we apply 4-learnable-region, 8-learnable-region and 16-learnable-region settings to DRConv for analysing the influence of different number of regions. 

The results of comparing our DRConv with standard convolution are shown in Table~\ref{table:det_and_seg}. From the results we can see that DRConv with 8 regions in DetNAS-300M can significantly improve the performance by $1.8\%$ for detection, only two DRConv layers with 16 regions in FPN of Mask R-CNN can improve the performance by $1.2\%$ for detection and $1.1\%$ for segmentation on COCO's standard AP metric. DRConv utilizes guided mask to divide spatial dimension into groups so that each filter can focus on special context. On other hand, the noise like background can be easily separated from other regions of interest and most of the filters can concentrate on important regions. For different number of shared regions, the results are shown that DRConv can achieve better performance when we divide spatial dimension to more regions. More divided regions make every group's context more dedicated and each filter can be optimized more easily.

\section{Ablation Study}
The ablation experiments are conducted on classification (ImageNet 2012~\cite{russakovsky2015imagenet}) and face recognition (MS1M-V2~\cite{guo2016ms}). The experimental settings are the same as that in section 4. In this section, we analyze the semantic information of learnable guided mask, the influence of different model size. The influence of different region number and different spatial size with respect to DRConv is analyzed in the \emph{supplementary material}.
\\
\\
{\bf Visualization of dynamic guided mask.}
In order to explore the mechanism of the learnable guided mask in our method, we visualize guided mask with $m=4$ for image on classification task and face recognition task respectively. Fig.~\ref{Fig:visual} shows that our method successfully assigns filters to regions with the same semantics. In other words, we have made it possible to learn that different regions use different filters according to image semantics, which is reasonable and effective. Due to more clear semantic representation, guided mask may automatically form less number of regions in deeper layer.

It needs to be noted that the guided mask is totally decided by the spatial information distribution so that one region might be connected points or discrete points. The points of a region in shallow layers tend to be discrete because the feature is more relevant to the detailed context of the input image. The points of a region in deep layers tend to be connected because the points have a bigger receptive field which is more relevant to the semantic information. 

{\bf Different model size.}
Besides the investigation above, we conduct the ablation study of DRConv's performance on different model size. On ImageNet dataset, we carry out experiments on [0.5$\times$, 1$\times$, 1.5$\times$, 2$\times$] of ShuffleNetV2 and [0.25$\times$, 0.5$\times$, 0.75$\times$, 1$\times$] of MobileNetV2 to analyse the effectiveness of our DRConv. From our experimental results shown in Fig.~\ref{Fig:Comparision_DRConv_vs_Shuffle_and_Mobile}, smaller models with DRConv will gain more bonus than larger models. Obviously, small models are of less input channels and filters in each layer and they can't extract enough feature for next layer. By replacing standard convolution with DRConv, small models will conspicuously improve their capability of modeling semantic information, resulting in better performance. 

\section{Conclusion}
In this paper, we propose a new convolution named Dynamic Region-Aware Convolution (DRConv), which is motivated by partial filter sharing in spatial domain and successfully maintains translation-invariance property. Therefore, our proposed DRConv can entirely become substitute of standard convolution in any existing networks. We design a small learnable module to predict the guided mask for instructing the filters' assignment, which guarantees similar feature in a region can match the same filter. Furthermore, we design filter generator module to produce the customized filters for each sample, which makes it possible that different inputs can use their own specialized filters. Comprehensive experiments on several different tasks have shown the effectiveness of our DRConv, which can outperform state-of-the-art and other superior manually designed methods in classification, face recognition, object detection and segmentation. And our experiments in ablation study manifest that learnable guided mask plays a key role in filter distribution for each sample, which can help to achieve better performance.

\newpage

{\small
\bibliographystyle{ieee_fullname}
\bibliography{egbib}
}

\end{document}